\documentclass{ieeetj}
\usepackage{cite}
\usepackage{amsmath,amssymb,amsfonts}
\usepackage{algorithmic}
\usepackage{graphicx,color}
\usepackage{textcomp}
\usepackage{xcolor}
\usepackage{hyperref}
\usepackage{cite}
\usepackage{xspace}
\usepackage{multirow}
\hypersetup{hidelinks=true}
\usepackage{colortbl}
\usepackage{algorithm,algorithmic}
\usepackage{subcaption}
\def\BibTeX{{\rm B\kern-.05em{\sc i\kern-.025em b}\kern-.08em
    T\kern-.1667em\lower.7ex\hbox{E}\kern-.125emX}}
\AtBeginDocument{\definecolor{tmlcncolor}{cmyk}{0.93,0.59,0.15,0.02}\definecolor{NavyBlue}{RGB}{0,86,125}}

\def\OJlogo{\vspace{-4pt}$<$Society logo(s) and publication title will appear here.$>$}
\def\seclogo{\vspace{10pt}$<$Society logo(s) and publication title will appear here.$>$}

\def\authorrefmark#1{\ensuremath{^{\textbf{#1}}}}

\newcommand{\Algoname}{KIPPS\xspace}

\begin{document}
\receiveddate{XX Month, XXXX}
\reviseddate{XX Month, XXXX}
\accepteddate{XX Month, XXXX}
\publisheddate{XX Month, XXXX}
\currentdate{XX Month, XXXX}
\doiinfo{XXXX.2022.1234567}

\markboth{\Algoname: Knowledge infusion in Privacy Preserving Synthetic Data Generation}{Kotal {et al.}}

\title{\Algoname: Knowledge infusion in Privacy Preserving Synthetic Data Generation}

\author{Anantaa Kotal\authorrefmark{1, 2}, and Anupam Joshi\authorrefmark{1}, Fellow, IEEE}
\affil{Department of C.S.E.E., University of Maryland at Baltimore County, Baltimore MD USA}
\affil{Department of Computer Science, The University of Texas at El Paso, El Paso TX USA}
\corresp{Corresponding author: Anantaa Kotal (email: akotal@utep.edu).}
\authornote{This paragraph of the first footnote will contain support information, including sponsor and financial support acknowledgment. For example, ``This work was supported in part by the U.S. Department of Commerce under Grant 123456.''}

\begin{abstract}
The rapid evolution of Machine Learning (ML) has led to a large number of applications across diverse domains. The success of ML models lies in the availability of large volumes of data to train on.  Since data is siloed in many areas like healthcare and cybersecurity, this requires data to be shared across sites to create a model that is not narrow and brittle. The necessity for sharing data raises significant questions on data privacy and confidentiality. In many domains, data sharing is regulated or outright prohibited, impacting the success of ML models. One approach to this problem is Federated learning, but it has its limitations. An alternate approach is synthetic data generation, which  is a scalable and practical solution for privacy-preserving data sharing. It allows organizations to collaborate on research and analysis without compromising sensitive information. Generative Deep Learning models have proven instrumental in addressing privacy concerns by generating synthetic data that retains statistical characteristics while concealing individual details. The integration of privacy measures, including differential privacy techniques, ensures a provable privacy guarantee for the synthetic data. However, challenges arise for Generative Deep Learning models when tasked with generating realistic data, especially in critical domains such as Cybersecurity and Healthcare. Generative Models optimized for continuous data struggle to model discrete and non-Gaussian features that have domain constraints. Challenges increase when the training datasets are limited and not diverse. In such cases, generative models create synthetic data that repeats sensitive features, which is a privacy risk. Moreover, generative models face difficulties comprehending attribute constraints in specialized domains. This leads to the generation of unrealistic data that impacts downstream accuracy. To address these issues, this paper proposes a novel model, \Algoname that infuses Domain and Regulatory Knowledge from Knowledge Graphs into Generative Deep Learning models for enhanced Privacy Preserving Synthetic data generation. The novel framework augments the training of generative models with supplementary context about attribute values and enforces domain constraints during training. This added guidance enhances the model's capacity to generate realistic and domain-compliant synthetic data. The proposed model is evaluated on real-world datasets, specifically in the domains of Cybersecurity and Healthcare, where domain constraints and rules add to the complexity of the data. Our experiments evaluate the privacy resilience and downstream accuracy of the model against benchmark methods, demonstrating its effectiveness in addressing the balance between privacy preservation and data accuracy in complex domains.
\end{abstract}

\begin{IEEEkeywords}
Enter key words or phrases in alphabetical order, separated by
commas. Using the \textit{IEEE Thesaurus} can help you find the best
standardized keywords to fit your article. Use the \underline{\href{https://www.ieee.org/publications/services/thesaurus.html}{thesaurus
access request form}} for free access to the \textit{IEEE Thesaurus}.
\end{IEEEkeywords}


\maketitle

\section{INTRODUCTION}
Data is a crucial asset in the information age, profoundly affecting many aspects of modern society. As technologies for data mining and data analytics  advance, the potential value of data become more apparent. Businesses and organizations recognized that data, when properly leveraged, could yield valuable insights, drive informed decision-making, and offer a competitive advantage. The rapid evolution of machine learning models further intensified the value and demand for data across diverse domains. Machine learning relies heavily on large and diverse datasets for training models effectively. More data allows these models to learn intricate patterns, improve accuracy, and generalize well to new situations. As machine learning applications continue to expand across various industries, the quality and quantity of data play a crucial role in the success of these models. 

While the availability of data is crucial, particularly in the context of machine learning, it also gives rise to potential risks to the privacy of individuals. Organizations, driven by the desire to extract valuable insights from vast datasets, often gather crucial but sensitive information. This practice has raised concerns on how individual privacy is compromised in the process. The recognition of these concerns has prompted a growing emphasis on data privacy, leading to the establishment of regulations like the General Data Protection Regulation (GDPR). These regulations sets forth guidelines and standards for the protection of personal data, making it a legal necessity for organizations to comply with the privacy rules. As the need for data privacy becomes more urgent, there is increasing effort to develop privacy-preserving measures for data sharing. Privacy-preserving techniques involve the implementation of methods that allow organizations to extract meaningful information from data without compromising the identity or sensitive details of individuals. These measures aim to strike a balance between retaining data utility and safeguarding privacy.

Synthetic data generation stands out as a crucial solution in the face of growing concerns about privacy. Rather than sharing raw, identifiable data, organizations can generate synthetic samples that retain the overall structure and patterns of the data. This allows collaborative research and analysis without the need to expose sensitive information. Generative Deep Learning has emerged as a powerful tool for enhancing privacy in various applications, addressing concerns related to data security and confidentiality. The application of Generative models, particularly in the context of privacy, revolves around its ability to generate synthetic data. By using generative models, like Generative Adversarial Networks (GANs), researchers and organizations can create synthetic datasets that maintain the statistical characteristics of the original data without revealing specific details about individuals \cite{brock2018large, isola2017image, , park2018data, han2018gan, karras2019style}. This facilitates the development and testing of machine learning models without compromising privacy. The incorporation of privacy measures, such as the integration of differential privacy techniques with Generative Deep Learning, offers an additional layer of protection and provides a provable privacy guarantee.

While Generative Deep Learning offers promising solutions for privacy preservation in the realm of images and text, it encounters challenges when tasked with generating realistic tabular data, such as for Cybersecurity, Healthcare etc \cite{jordon2018pate, xu2018synthesizing}. Generative Adversarial Networks (GANs) are designed for continuous Gaussian distributions and encounter difficulties when faced with discrete features and non-gaussian features. Most datasets in domains such as Cybersecurity and Healthcare contain a mix of discrete and non-gaussian features, posing difficulties for these models. The challenge is amplified due to the lack of diverse datasets for training Generative Models. Generative models, trained on limited data, have visibility of only a subset of the possible feature values. Consequently, during synthesis, they can only produce those limited values. As a result, the synthetic data is less diverse and poses a privacy risk as it contains repetitions of features that may be sensitive. Furthermore, the possible values of discrete features in such datasets is often very high. This increases the complexity of generative models which is directly proportional to the range size of discrete values. Consider the example of network activity data generation, where one of the significant attributes is IP address. The IP address in the training data is limited and the entire actual range of IP address is enormous. The model training on the limited data capture has narrow scope of view and the complexity of modeling the entire range of values is very high and infeasible. 

Moreover, the scarcity of data poses a challenge for generative models in comprehending attribute constraints imposed by the domain. Datasets in specialised domains often contain specific meanings and relationships between attribute values, which are challenging for these models to deduce from limited training data. For instance, in network activity data, specific protocols correlate with particular port numbers, and an incorrect pairing is not just improbable but inaccurate. Generative models may fail to explicitly capture these constraints, leading to the generation of unrealistic data that is distinguishable from the original. This discrepancy can misguide downstream tasks, like classifiers distinguishing between legitimate and anomalous behavior. Going back to the example of network activity data, IP addresses have special meanings and associated constraints. For example, a DNS lookup request can only go from the Home IP to the Gateway Server IP, never the otherway round. This information is not explicit to the generative model from the training data.

To address these challenges, it is essential to provide the generative models with added context about the domain. The domain knowledge provides additional information attribute values and ranges, and awareness of domain-specific patterns or constraints. This domain specific context enhances the generative models ability to interpret data meaningfully. In this study, we introduce a novel framework, \Algoname, for infusing knowledge into privacy-preserving generative models during training, enhancing their capacity to generate realistic and domain-compliant synthetic data. Our findings demonstrate that incorporating additional domain-specific context and enforcing constraints during training improves the generative model's learning process, resulting in synthetic datasets that are better representative of the original data in downstream tasks. We demonstrate the efficacy of our model by generating synthetic substitutes of real world tabular datasets in different domains, including Cybersecurity ad Healthcare. We show that the synthetic dataset retains significant information from the dataset and able to replace the original data is downstream tasks. Furthermore, we show that the \Algoname model is resilient against state-of-the-art privacy attacks. The \Algoname model offers a practical solution to the data sharing problem, balancing data utility and privacy resilience.

\section{Background}
\label{sec_background}
\subsection{Data Synthesis for Privacy}
In addressing data privacy challenges, strategies like data anonymization, secure cryptographic methods, and distributed model release have been used, but each has limitations. Data anonymization, a common technique, can be compromised with future information, and secure cryptographic methods restrict open access to data \cite{sweeney2002k, machanavajjhala2007diversity, dwork2006calibrating, rocher2019estimating}. Synthetic data generation emerges as a promising solution, offering a practical way to share data while preserving privacy. This approach creates entirely new datasets without real individual associations. In critical fields like healthcare and security, where micro-data is essential, synthetic data generation ensures reliable conclusions.  These methods try to maintain the inherent properties of the original dataset, thus enabling the use of synthetic data as replacement in downstream tasks with minimal cost to accuracy. This provides a cost-effective, scalable alternative that encourages data sharing, reduces privacy costs, and enables collaboration. 

Generative Adversarial Networks (GANs) are a powerful class of Generative Deep Learning models widely recognized for their success in creating synthetic data closely resembling the original dataset. GANs excel in generating high-fidelity synthetic data, particularly in image and text domains \cite{brock2018large, isola2017image, zhang2017stackgan, wang2018high}.  In real-world domains like Cybersecurity and Healthcare, traditional GANs encounter challenges. The data in these domains is typically tabular, containing both discrete and continuous values. Generating realistic tabular data poses difficulties in modeling complex relationships and diverse value ranges, rendering GANs less effective compared to their success with continuous and image-based datasets. In the 2019 paper by Xu et al. \cite{xu2019modeling}, the author proposes a GAN model to address the challenges with tabular data. Specifically, they address the issues of multi-modality and non-gaussian nature of continuous variables and class imbalance in tabular data. In the 2021 paper by Kotal et al.\cite{kotal2022privetab}, further used this model to create a framework for privacy preserving data generation in tabular data. The authors enforce t-closeness in the generated data to the original dataset to preserve privacy.

However, generative models are not sufficient by itself in guaranteeing privacy. We need strong mathematical foundation for the privacy guaranteed through this process. Differential Privacy provides provable privacy guarantee for a randomized algorithm. As a practical and implementable method of differential privacy Dwork et al. \cite{dwork2006calibrating} propose noise addition to true output. Differential privacy provides the mathematical foundation for ensuring privacy in any algorithmic setting. Differential Privacy Stochastic Gradient Descent (DP-SGD), introduced by Abadi et al.\cite{abadi2016deep}, is a privacy-preserving deep learning algorithm that blends stochastic gradient descent (SGD) principles with differential privacy. The NIST-winning Probability Graphical Models framework \cite{mckenna2019graphical} utilizes a graphical model assumption for data distribution, employing parametric and non-parametric techniques to estimate parameters maximizing entropy. PrivBayes \cite{zhang2017privbayes} is a privacy-preserving data generation method using Bayesian networks with binary nodes. Abay et al. \cite{abay2019privacy} propose an autoencoder-based data synthesis method, partitioning a sensitive dataset into label-based groups. Building on Abadi et al.'s work \cite{abadi2016deep}, various GAN models have been proposed that combine Differential Privacy Stochastic Gradient Descent (DP-SGD) with Generative Adversarial Networks (GANs) to generate differentially private synthetic data \cite{xie2018differentially, torkzadehmahani2019dp}. These models introduce noise into the GAN's discriminator during the training process to enforce differential privacy. The crucial aspect of DP's guarantee of post-processing privacy means that by safeguarding the GAN's discriminator, it ensures differential privacy for the parameters of the GAN's generator.

However, GANs continue to face challenges, specifically in generating synthetic tabular data for domains with limited training data availability leading and strict domain constraints. Limited view of discrete value attributes, complexity of attribute ranges and conditional constraints of the attributes make it harder for the GANs to model the data accurately and pose significant privacy risks. We discuss the challenges with privacy preserving synthetic data generation using GAN further in Section \ref{sec_framework}. We propose that domain knowledge can provide the added context needed to train the generative model for alleviating these challenges.

\subsection{Knowledge Guided Learning} 
Generative Adversarial Networks (GANs) trained solely on observed tabular data encounter limitations in comprehending intricate attribute relationships, such as the interplay between categorical variables or the constraints governing certain feature combinations. For instance, in tabular datasets, specific attributes may adhere to predefined rules or correlations, such as the acceptable range of values for categorical variables or the permissible combinations of features. Without explicit guidance, GANs lack the inherent ability to conform to these rules, potentially leading to the generation of synthetic data that deviates from the established constraints.

In domains where data adherence to predefined rules is paramount, such as Cybersecurity and Healthcare, incorporating domain knowledge becomes indispensable. By leveraging knowledge guidance techniques, we can explicitly convey these constraints to the generative model. This process equips the model with the necessary contextual understanding to generate synthetic data that adheres to the established rules and preserves the realism and integrity required for accurate analysis within the observed system.

Knowledge Graphs (KG) serve as a versatile graph-structured data model designed for knowledge representation and reasoning. Within a KG, information is organized into semantic triples, consisting of a subject, predicate, and object. Subjects and objects correspond to nodes or entities in the graph, while predicates denote the functional relationships between them. The graph-like structure of KGs allows for the efficient storage of extensive information, and the network can be expanded with new knowledge as required. KGs are equipped with powerful reasoning capabilities, enabling the imposition of constraints on entities and the deduction of new knowledge. For instance, specifying constraints such as the source IPs must not belong to a subnet or must originate from a specific external CIDR range. Leveraging these robust knowledge representation and reasoning abilities, KGs can enhance and enrich the data generation process.

Knowledge graphs excel at storing contextual information crucial for enhancing learning in distributed systems. The Unified Cybersecurity Ontology (UCO) stands out as a comprehensive ontology designed for cyber situational awareness in cybersecurity systems. Its integration has been demonstrated to significantly improve contextual awareness in machine learning (ML) systems, as evidenced by studies such as those by Piplai et al. \cite{piplai2020creating} and Narayanan et al. \cite{narayanan2018early}.

In a related context, Hui et al. \cite{hui2022knowledge} have introduced a knowledge-enhanced Generative Adversarial Network (GAN) designed to generate Internet of Things (IoT) traffic data for devices from various manufacturers. Their approach however is specific to IoT traffic and uses knowledge injection to set the conditions for IoT traffic generation. In this paper, we inject knowledge about network traffic into the training of the GAN by adding the Knowledge base as an independent discriminator. 

\section{Objective and Proposed Method}
\label{sec_framework}
Generative models have shown remarkable success in duplicating datasets, yet challenges persist when it comes to generating synthetic tabular data. This discussion explores three main problems faced by generative models in this context.
\begin{enumerate}
    \item  \textbf{Limited diversity in training data}  One significant challenge is the limited availability of data in the training set, offering an incomplete view of attributes. This is particularly problematic for discrete attributes where the training data may only cover a subset of possible values. Consider a practical scenario in a Network activity dataset, where the available IP addresses are both limited and private. This restricted view excludes a broader spectrum of valid IPs. Generative deep learning models, when trained on such data, tend to replicate only the discrete values observed during training. This is a privacy risk as the specific value of the IPs are considered sensitive. In spite of the privacy preserving measure, the risk of re-identification is not removed as the sensitive attribute value is repeated in the synthetic data. To truly anonymize the dataset, we need to allow the generative model to consider other possible values of the sensitive attribute.

    \item \textbf{Complexity of Discrete Attributes} Generative models for tabular data categorize attributes as either continuous or discrete. The complexity of the generative model is proportional to the range of discrete attribute values. In real world scenarios, the set of discrete values is large. The high input cardinality of discrete attributes increases the complexity of generative models and makes them intractable. In network activity data, for example, port number can be any valid number represented by 2 bytes. Hence, to consider all possible port values, the complexity of the learning model increases by the order $2^16$. For multiple attributes, the order of complexity increases by the product of number of attributes multiplied by their range. In practical situations, it is infeasible to train generative models for multiple discrete values with high cardinality. Added context about the domain can help us alleviate this problem by grouping attribute by property and reducing the complexity of the attribute.
    
    \item \textbf{Domain Constraints} In specialised domains, such as Cybersecurity, there are domain specific rules associated with the attributes. These rules are not apparent just from the training data. Consider the example of network activity data, where specific protocols and events are associated with specific port numbers. An unmatched port number with a network event is not just unlikely, but wrong. This distinction is not explicit to the generative model from the training data, thus they struggle to learn these complex dependencies. The result is unrealistic generated data that can easily be discerned from the original data and may mislead downstream tasks such as classifiers that are trying to separate legitimate traffic from attacks. Domain knowledge plays a crucial role in understanding these inherent rules between attributes. This information is explicitly stated in domain rules can provide the added guidance needed to train complex deep learning models.Injecting domain knowledge directly into the GAN training process accelerates convergence and improves the generation of realistic samples by codifying these restrictions.
    
\end{enumerate}
\begin{figure*}[t]
    \centering
    \includegraphics [width=\textwidth] {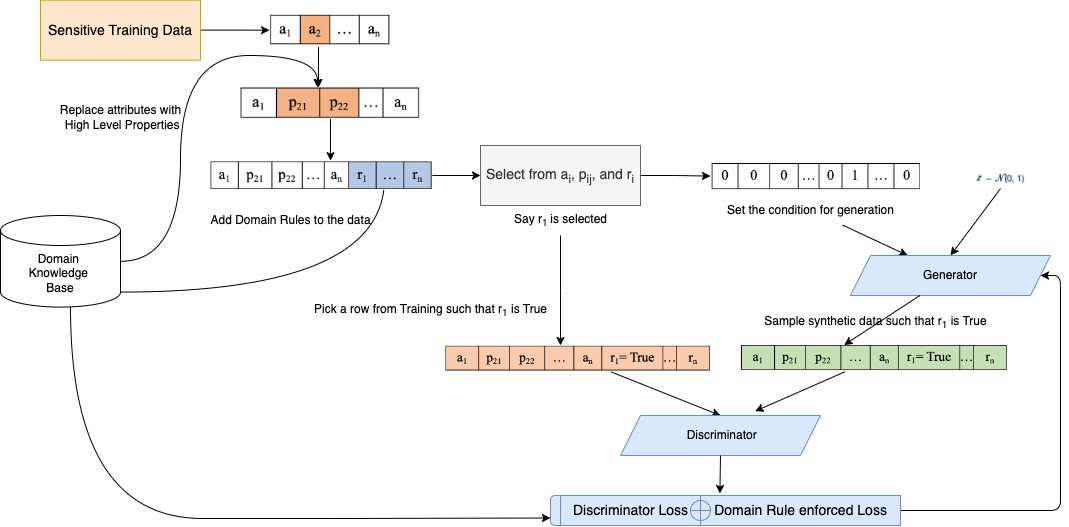} 
    \caption{\Algoname model}
    \label{fig_overall_arch}
\end{figure*}

To generate realistic privacy preserving synthetic data, while addressing the problems of limited diversity of sensitive attributes, complexity of discrete attributes, and domain restrictions, we propose a framework for Knowledge Infusion in Privacy Preserving Synthetic Data Generation (\Algoname). The novel framework introduces knowledge guidance from a domain Knowledge Graph (KG) into the training of a Generative Adversarial Network (GAN) and a conditional  Generative Adversarial Network (GAN) to address the problems of class imbalance and cross-correlation in attributes. The overall architecture for our proposed framework is illustrated in Figure \ref{fig_overall_arch}. The model is designed to satisfy 3 key requirements: 1. Provide domain context to limited and complex attributes 2. Enforce domain constraints in generated data 3. Satisfy Privacy Constraints in Synthetic Data. To satisfy these objectives, we break the \Algoname training into three phases:
\begin{enumerate}
    \item \textbf{Adding Domain Context to Training Data}
    \item  \textbf{Conditional Training with Domain Rule Enforced Loss}
    \item  \textbf{Differentially Private Discriminator}
\end{enumerate}
In this section, we describe the details of our framework and how it works to alleviate the key challenges identified with domain restricted synthetic tabular data generation.

\subsection{Adding Domain Context to Training Data}
\subsubsection{Replace Attributes with Domain Properties} Insufficient data availability in the training set, especially for discrete attributes, results in an incomplete representation of possible attribute values. Generative deep learning models trained on such limited data may reproduce only the observed discrete values, posing a privacy risk. Domain knowledge about attribute ranges and properties can help us overcome this. For sensitive attributes with limited data, we suggest replacing the specific values with general properties of the domain. This helps the generative model learn patterns that are true for attributes with similar properties, instead of just copying specific values. During generation, the model uses this knowledge to create new attributes with those shared properties, avoiding the issue of repeating the same values seen during training. For instance, in network data, if the list of IP addresses is small and private, using domain knowledge lets us understand other important properties like geo-location or organization. By replacing IP addresses with these properties, the generative model learns about the data and creates new IP addresses that share these important properties, even if they were not in the training data.

\subsubsection{Grouping Attributes by Properties}
In addition to overcoming the challenges of diversity, domain properties can also help us simplify the complexity of discrete attributes. The specific values of discrete attributes are many and often infeasible to model. However, it is critical to remember that all possible values of discrete attributes do not need to be treated independently. Using domain knowledge, to group attribute values by their properties can reduce dimensionality of input to the training model. This significantly reduce the computational burden on the model. For example, in network activity data, port numbers can range from 0 to 65535. It is infeasible to uniquely represent each possible port number at training. However, domain knowledge informs us that we do not need unique representation for all possible port numbers. It is well known that any port between 49152 to 65535 fall under the group of dynamically assigned ports. If a network event requires a dynamically assigned port, it is irrelevant which specific port value is used. Hence, instead of having a unique representation of all possible port values, we can group them together by their properties.

\subsubsection{Conditional Rules} 
Having a small amount of training data means the model doesn't see the full picture of the domain, making it difficult or even impossible for the model to learn specific conditions in the data just from the training set. However, we can overcome this limitation by incorporating domain knowledge, which helps identify the rules triggered at each data point. By supplementing the training data with this knowledge, we enhance the model's ability to understand and apply conditional restrictions in the dataset. For example, in network activity data, a DNS lookup request triggers a specific rule, where the request goes from the communicating device to the gateway server. Furthermore, the request has to be followed by a legitimate response from the server subsequently. The source and destination IP address for the DNS lookup request is predetermined by the domain rule. Hence, it is important that the generative model not only learn the association between the two IP addresses but the associated rule. It is thus helpful to explicitly add the context of the triggered rule to the training data.


\subsection{Conditional Training with Domain Rule Enforced Loss}
\subsubsection{Conditional Training}
To represent the Tabular Data as an input to the generative model, the discrete attributes are represented as one hot vectors,  $\{d_{1,j},...,d_{N_d,j}\}$ where $d_{i,j}$ is one hot representation of the j-th value of the i-th discrete attribute. The discrete attributes that were masked by their broader properties in the previous step are represented as the one hot vectors of their feature mask, where $f_{i,j}$ is one hot representation of the j-th value of the i-th .feature mask We follow the CTGAN method \cite{xu2018synthesizing} of representing continuous values through mode-specific normalization. Continous attributes are represented as one hot representation of the selected mode $\beta_{i,j}$ and scalar representation of value within mode $\alpha_{i,j}$. The conditional rules are added as one hot vectors of binary flags $\{r_1,...,r_k\}$. The representation of a single row is thus a concatenation of continuous and discrete variables:
\begin{equation}
    row_i = \alpha_{i,j} \oplus \beta_{i,j} \oplus ... \oplus d_{i,j} \oplus ... \oplus f_{i,j} \oplus ... \oplus r_i
\end{equation}

When training we have to ensure that all classes are sufficiently represented in training. This is enforced by probability based conditional vector input to the generator. The \textit{cond} vector is one hot vector where only one of $d_{i,j}$, $f_{i,j}$ or $r_{i}$ is set as 1 for all $i, j$. For example, if $r_{i} = 1$, denoting the condition that the \textit{i}-th rule is \textit{True}, the outcome of the Generator is $G(z|r_{i} = True)$. The input to the Critic is $G(z|r_{i} = True)$ and a row $row_i$ from the training data such that it satisfies $r_{i} = True$. The critic gives a scored outcome of the distinction between $G(z|r_{i} = True)$ and $row_i$. 

\subsubsection{Domain Rule enforced Generator Loss} 
The model is trained using the WGAN loss with gradient penalty \cite{gulrajani2017improved}. The critic loss is defined as, 
\begin{equation}
L_C={\underset{\tilde{\boldsymbol{x}} \sim \mathbb{P}_g}{\mathbb{E}}[D(\tilde{\boldsymbol{x}})]-\underset{\boldsymbol{x} \sim \mathbb{P}_r}{\mathbb{E}}[D(\boldsymbol{x})]}+{\lambda \underset{\hat{\boldsymbol{x}} \sim \mathbb{P}_{\hat{\boldsymbol{x}}}}{\mathbb{E}}\left[\left(\left\|\nabla_{\hat{\boldsymbol{x}}} D(\hat{\boldsymbol{x}})\right\|_2-1\right)^2\right]}
\end{equation}

The generator loss is defined as,
\begin{equation}
L_G=-{\underset{\tilde{\boldsymbol{x}} \sim \mathbb{P}_g}{\mathbb{E}}[D(\tilde{\boldsymbol{x}})]}
\label{generator_loss}
\end{equation}

However during training, the Generator is free to choose any set of one hot vector in its output. As we discussed, this goes against the explicit rules of the domain. For example, let there be a rule $r_k$, such that when $r_k$ is \textit{True}, the \textit{i}-th discrete vector must have the \textit{j}-th value and the \textit{i}-th feature mask must be the \textit{j}-th value. However, there is nothing in the Generator that forces it to generate  $d_{i,j}=1$ and $f_{i,j}=1$, when $r_k = 1$. The mechanism proposed to enforce the condition $(d_{i,j}=1 \land f_{i,j}=1 \land r_k = 1)$ is to penalize the Generator loss by adding the cross-entropy, \textbf{\textit{H}}, between the domain-enforced condition $(d_{i,j}=1 \land f_{i,j}=1 \land r_k = 1)$ and the generator output, averaged over all the instances of the batch. Hence, the Generator loss in Equation \ref{generator_loss} is modified to include the rule based cross entropy loss:

\begin{equation}
L_G=\underbrace{-{\underset{\tilde{\boldsymbol{x}} \sim \mathbb{P}_g}{\mathbb{E}}[D(\tilde{\boldsymbol{x}})]}}_{\text {Original Generator loss}} + \underbrace{H(KG(cond.))}_{\text {Domain Rule enforced Loss}},
\end{equation}

where, $KG(cond.)$ is domain rule extracted for the conditional vector \textit{cond}.

\subsection{Differentially Private Discriminator}
To account for privacy in the \Algoname model, we need to incorporate the differential private framework. We employ the privacy accountant as described in the seminal work by Abadi et al. \cite{abadi2016deep} to track the privacy loss. To make the model differentially private, we need to add noise within the differential privacy paradigm to the output of our model. Fang et al. \cite{fang2022dp} postulate that it is sufficient to add random perturbation only in the training of the discriminator. Since only the discriminator has access to the real data, it will be sufficient to control the privacy loss. This approach reduces the challenges of achieving convergence with differential privacy and the complexity of estimating privacy loss.

We achieve differential privacy in the conditional discriminator through the DP-SGD \cite{abadi2016deep} method. At each step of the stochastic gradient descent (SGD), we compute the gradient for a random subset of examples, as:
\newline For each $i \in L_t$, compute $\mathbf{g}_t\left(x_i\right) \leftarrow \nabla_{\theta_t} \mathcal{L}\left(\theta_t, x_i\right)$ 

The we clip the $l_2$ norm of each gradient, compute the average. 
$$
\overline{\mathbf{g}}_t\left(x_i\right) \leftarrow \mathbf{g}_t\left(x_i\right) / \max \left(1, \frac{\left\|\mathbf{g}_t\left(x_i\right)\right\|_2}{C}\right)
$$

Finally we add noise in order to
protect privacy
$$
\tilde{\mathbf{g}}_t \leftarrow \frac{1}{L}\left(\sum_i \overline{\mathbf{g}}_t\left(x_i\right)+\mathcal{N}\left(0, \sigma^2 C^2 \mathbf{I}\right)\right)
$$

\section{Experimental Method}
\label{sec_datasets}
To evaluate the efficacy of the \Algoname model, we tested it on datasets from various domains, ensuring its versatility and robustness across different types of data. For the healthcare domain, we utilized the Opioid Dataset and CDC Diabetes Health Indicator dataset, which include comprehensive patient records with attributes such as demographics, vital signs, and medical history. For socio-economic studies, we employed the Adult Income dataset from the UCI Machine Learning Repository, which features demographic and income-related attributes. We also tested the model on the Network Traffic dataset and the UNSW-NB15 dataset, which include detailed records of network activity with attributes like source and destination IPs, packet sizes, and timestamps. These datasets are crucial for cybersecurity analysis and intrusion detection, containing millions of instances with various attributes representing different network traffic features. 

Each of these datasets presented unique challenges, such as varying degrees of attribute complexity, domain-specific constraints, and the need for privacy preservation. By incorporating domain knowledge from Knowledge Graphs tailored to each dataset, \Algoname effectively generated realistic, privacy-preserving synthetic data that adhered to domain-specific constraints and relationships. This comprehensive evaluation across multiple domains demonstrated the model's ability to generalize and perform effectively in diverse scenarios, highlighting its potential for wide-ranging applications. In this section, we provide a description of the datasets, the construction of knowledge bases and the experimental setup for evaluating the \Algoname model.

\subsection{Adult Income Dataset:} The Adult Income dataset \cite{} from the UCI Machine Learning Repository comprises 48,842 instances with 14 attributes, including age, education, occupation, and income. The dataset is used to predict whether an individual's income exceeds $\$50,000$ per year based on demographic and employment information. Attributes include both continuous variables (like age and hours-per-week) and categorical variables (like workclass and marital status). The rules and constraints for the Adult Income dataset include that individuals with advanced degrees (e.g., Bachelors, Masters) are more likely to earn over $\$50,000$ per year. Certain occupations, such as executive or professional roles, typically require higher education levels and correlate with higher incomes. Furthermore, higher education levels have a minimum number of years spent in education. These constraints are added to the Adult Income Knowledge Base and provides additional context to the \Algoname model. 

\subsection{Opioid Prescription Dataset:} The Opioid Prescription Dataset \cite{} focuses on characterizing opioid prescribing patterns among uninsured patients seeking emergency medical care. The study aimed to build predictive machine learning models to understand prescribing trends based on demographic factors. Uninsured patients were found to be less likely to receive opioid medication but more likely to receive non-opioid alternatives and less likely to receive antimicrobial prescriptions. Key contributing factors included housing status, comorbidities, and recidivism. The analysis was conducted on 68,969 emergency department visits between January 2017 and December 2018, revealing notable differences in prescribing patterns between uninsured and insured patients. For this study, the knowledge base was constructed by integrating prescription rules established by the University of Maryland Baltimore (UMB) Emergency Medicine Department. These rules provided guidelines for opioid prescriptions in emergency medical settings. By incorporating these rules into the knowledge base, the study ensured that the predictive machine learning models aligned with the prescribing practices endorsed by the UMB Emergency Medicine Department.

\subsection{CDC Diabetes Health Indicator:} The CDC Diabetes Health Indicator dataset contains health-related data aimed at assessing the prevalence and management of diabetes across different populations. It includes attributes such as age, gender, body mass index (BMI), blood pressure, and glucose levels, among others. The dataset facilitates analyses to understand the prevalence of diabetes and its associated risk factors, aiding in public health interventions and policy decisions. The rules and constraints for the CDC Diabetes Health Indicator dataset are based on established medical guidelines and epidemiological observations. These guidelines dictate thresholds for blood glucose levels, BMI ranges, and blood pressure values indicative of diabetes risk. Additionally, demographic factors such as age and gender may influence diabetes prevalence rates, guiding analyses to account for population characteristics when assessing diabetes trends.

\subsection{Network Traffic Data:} The Network Traffic dataset, obtained from a previous research on Network Traffic simulation in digital twins, encompasses network traffic data from an experimental IoT setup. This setup includes diverse IoT devices like a Blink camera, a smart plug with a lamp, and a motion sensor. Using network capture tools like Wireshark, the study monitored the communication patterns within this IoT ecosystem, focusing on events such as motion detection by the Blink camera, lamp activation by the smart plug, and communications involving the motion sensor and tag manager. Essential features such as Source IP address, Destination IP Address, Source Port, Destination Port, and Protocol were captured, allowing for analysis of typical device communications. Additionally, deliberate attack scenarios were simulated during data collection to generate instances for training Network Intrusion Detection Systems (NIDS), enabling the identification of anomalous behaviors and pre-emptive detection of attacks. The Network KG \cite{} was designed to define concepts in network activity data, extending from the UCO ontology. Key entities include "networkEvent", "networkEventProperty", and "domainURL", describing individual network events, their properties (such as Protocol, sourceIP, destinationIP, sourcePort, and destinationPort), and domain URLs. Network events have specific ranges of IPV4 and IPV6 addresses and port range values are also restricted. The Knowledge Graph reasoner can be utilized to return queries of valid IP, port, and protocol combinations for a given event type, providing true or false constants against the query. The Network KG provides additional context to the \Algoname model about Network Traffic Data.

\subsection{UNSW-NB15 Data: }
The UNSW-NB15 dataset's raw network packets were crafted in the UNSW Canberra Cyber Range Lab using the IXIA PerfectStorm tool. This process aimed to create a blend of genuine contemporary normal activities and synthetic modern attack behaviors. The tcpdump tool was employed to capture 100 GB of the raw traffic in Pcap files. This dataset encompasses nine attack types, including Fuzzers, Analysis, Backdoors, DoS, Exploits, Generic, Reconnaissance, Shellcode, and Worms. Utilizing tools such as Argus and Bro-IDS, along with the development of twelve algorithms, resulted in the generation of 49 features along with their corresponding class labels.

\section{Evaluation Metrics}
One of the primary utility of tabular datasets is to support various downstream machine learning tasks across different domains. Whether it's in healthcare, cybersecurity, finance, or other fields, machine learning models often rely on high-quality data for effective training. However, obtaining reliable training data remains a challenge in many domains due to privacy concerns and data availability issues. Privacy-preserving generative models offer a promising solution by generating synthetic datasets that maintain the statistical properties of the original data while preserving privacy. 

To evaluate the effectiveness of privacy preserving frameworks of data sharing, we look at three types of metrics.
\begin{enumerate}
    \item \textbf{Fidelity Metrics: } to give a broad sense of how “close” the synthetic data are to the original data.
    \item \textbf{Utility Metrics: } to show the effectiveness of the synthetic data in downstream tasks, in this case, machine learning tasks. 
    \item \textbf{Privacy Metrics: } to show the resilience of the synthetic data generating models against state-of-the-art privacy attacks. 
\end{enumerate}

\subsection{Fidelity Metrics}
\subsubsection{\textbf{Propensity Score Accuracy}} 
\label{subsec_pmse}
The score, Predictive Mean Squared Error (PMSE), is calculated by comparing the predicted propensity scores of the synthetic dataset to the expected probability of being in the synthetic group. Propensity scores, estimated using classification and regression trees (CART) models trained on a combined dataset of original and synthetic data, are used to predict group membership. The PMSE is computed as the mean squared error between these predicted propensity scores and the expected probability of being in the synthetic group. Essentially, it measures the accuracy of propensity score estimation from the synthetic data in predicting group membership. Lower PMSE values indicate better predictive accuracy and, consequently, higher utility of the synthetic data. By comparing PMSE values across different synthetic data generation methods or parameter settings, we can evaluate the effectiveness of each method in preserving the distributional characteristics and utility of the original data. We use the approach as proposed by Snoke et al. \cite{snoke2018general} to calculate the PMSE ratio.

\subsubsection{\textbf{Distributional Distance}}
\label{subsec_chi}
The evaluation of synthetic data quality involves assessing accuracy on specific data elements, such as univariate differences between synthetic and original data or disparities in regression analysis estimates. Instead of employing direct distance measures, which may not be suitable for datasets with varying sizes, we utilize distributional distance metrics. Specifically, we employ the Chi-square test $(\chi^2)$ for categorical variables and the Kolmogorov-Smirnov (KS) test for continuous variables, as proposed Bowen et al \cite{bowen2019comparative}. These tests measure distributional dissimilarities and accommodate datasets of differing sizes, allowing for comparison across various synthetic datasets. To address the scale differences among variables, we average the univariate distances across the entire dataset after converting them to p-values. Although these p-values are not used in traditional null hypothesis significance testing, they serve as scale-free distance measures. The resulting utility metric is the average of the p-values for each variable, with categorical and continuous variables treated separately to account for potential differences in the synthesis processes for each type of variable.

\subsubsection{\textbf{Regression-Based Evaluation Metrics}} 
\label{subsec_regr}
The score is calculated by conducting regression analysis on both original and synthetic datasets, followed by the computation of two specific utility metrics for each coefficient in the models, as used by Woo et al. \cite{woo2015generalised}
    \begin{enumerate}
        \item \textbf{Confidence Interval Overlap Measure:} This metric quantifies the average overlap between confidence intervals estimated from the original and synthetic data for a single estimate. It assesses how much the confidence intervals estimated for the original and synthetic data overlap, providing insights into the consistency of estimation between the two datasets.
        \item \textbf{Standardized Difference in Coefficient Values:} This metric evaluates how far the coefficients derived from synthetic data deviate from those of the original data, regardless of the confidence interval width. It focuses on the magnitude of differences in coefficient values, providing a measure of the extent to which synthetic data accurately represents the original dataset in terms of regression coefficients.
    \end{enumerate}
These metrics aid in evaluating the quality of synthetic data by providing insights into the consistency and accuracy of regression model estimates between original and synthetic datasets. By comparing these metrics across different coefficients, researchers can assess the overall performance of synthetic data generation methods in preserving the inferential characteristics of the original data.

\subsection{Utility Metrics:}
The \textbf{Classifier Performance Utility Metric} evaluates the effectiveness of synthetic data by comparing the performance of various machine learning classifiers trained on synthetic data against those trained on original data. The process involves preparing two separate training datasets, one from the original data and one from the synthetic data, and using a common test dataset sampled from the original data to ensure consistent evaluation. Several classifiers, including Random Forest (RF), K-Nearest Neighbors (KNN), Decision Tree (DT), Support Vector Machine (SVM), and Multi-Layer Perceptron (MLP), are trained separately on both the original and synthetic training datasets. Each classifier is then tested on the same original test dataset, and their performance is measured, typically using accuracy. The accuracy of classifiers trained on synthetic data is compared to the accuracy of those trained on original data for each type of classifier. The differences in accuracy are computed, and the average and standard deviation of these differences provide an overall assessment of the synthetic data's utility. This metric is crucial for evaluating synthetic data quality because it directly measures how well the synthetic data can support downstream machine learning tasks. High utility is indicated by minimal differences in performance, suggesting that the synthetic data effectively captures the relevant patterns of the original data, making it a viable substitute for training machine learning models. By testing across various classifiers, this metric ensures robustness, demonstrating that the synthetic data's utility is broadly applicable across different machine learning algorithms.

\subsection{Privacy Metrics}
\subsubsection{\textbf{Membership Inference Attack}} Membership inference attacks aim to determine whether a particular data point was included in the training dataset of a machine learning model, posing significant privacy risks \cite{hu2022membership}. To evaluate privacy resilience, a membership inference attack is performed on models trained with synthetic data. The process involves training a machine learning model on synthetic data and then attempting to infer the membership of data points that were part of the original dataset. Attackers use shadow models, which are trained to mimic the behavior of the target model, to generate predictions and construct an attack model that predicts membership status. The accuracy of this attack model indicates the likelihood of successfully inferring membership. The privacy resilience metric is quantified by measuring the attack model's overall accuracy. Low accuracy values suggest that the synthetic data provides strong privacy protection, making it difficult for attackers to accurately determine membership. This metric is crucial for evaluating synthetic data as it directly measures its ability to protect individual data points from being exposed through inference attacks, ensuring that the synthetic data can be safely used without compromising the privacy of individuals in the original dataset.

\subsubsection{\textbf{Attribute Inference Attack}} Attribute inference attacks aim to deduce unknown attributes of individuals based on other known attributes, posing significant privacy risks. To measure privacy resilience, an attribute inference attack is conducted on models trained with synthetic data. The process involves training a machine learning model on the synthetic dataset and using this model to predict missing or sensitive attributes of individuals in the original dataset \cite{annamalai2023linear}. The attack's success is quantified by the accuracy of these predictions. High prediction accuracy indicates that the synthetic data may reveal sensitive information, compromising privacy. Conversely, low accuracy suggests stronger privacy protection, meaning the synthetic data effectively conceals sensitive attributes from being inferred. This metric is crucial for evaluating synthetic data as it directly measures its ability to obscure individual attribute information, ensuring that the synthetic data can be used safely without revealing sensitive details about individuals in the original dataset. By focusing on the accuracy of attribute inference attacks, this metric provides a clear and quantifiable measure of how well the synthetic data protects against adversarial attempts to uncover hidden attributes.

\begin{table}[]
\resizebox{\columnwidth}{!}{%
\begin{tabular}{l|c|c|c|c|c|}
\cline{2-6}
                                       & \textbf{PMSE}                         & \textbf{$\chi^2$}            & \textbf{KS}                           & \textbf{CI Overlap}                    & \textbf{Std. Diff.}          \\ \hline
\multicolumn{1}{|l|}{\textbf{DPCTGAN}} & 0.24                                  & 0                            & 0.02                                  & 5.593                                  & \textbf{0.38}                \\ \hline
\multicolumn{1}{|l|}{\textbf{DPWGAN}}  & 0.24                                  & 0.02                         & 0.01                                  & 2.108                                  & 0.5                          \\ \hline
\multicolumn{1}{|l|}{\textbf{PATEGAN}} & 0.25                                  & 0.04                         & 0.02                                  & 5.688                                  & 0.5                          \\ \hline
\multicolumn{1}{|l|}{\textbf{TVAE}}    & 0.25                                  & \textbf{0.09}                & 0.06                                  & 1.249                                  & 9.74                         \\ \hline
\multicolumn{1}{|l|}{\textbf{KIPPS}}   & \cellcolor[HTML]{9B9B9B}\textbf{0.24} & \cellcolor[HTML]{9B9B9B}0.08 & \cellcolor[HTML]{9B9B9B}\textbf{0.06} & \cellcolor[HTML]{9B9B9B}\textbf{5.688} & \cellcolor[HTML]{9B9B9B}0.45 \\ \hline
\end{tabular}%
}
\caption{Comparison of PMSE, Distance and Regression based Evaluation metrics for Privacy Preserving Synthetic Data}
\label{tab:fidelity}
\end{table}

\begin{figure*}[]
\centering
\begin{subfigure}{0.49\textwidth}
  \includegraphics[width=\textwidth]{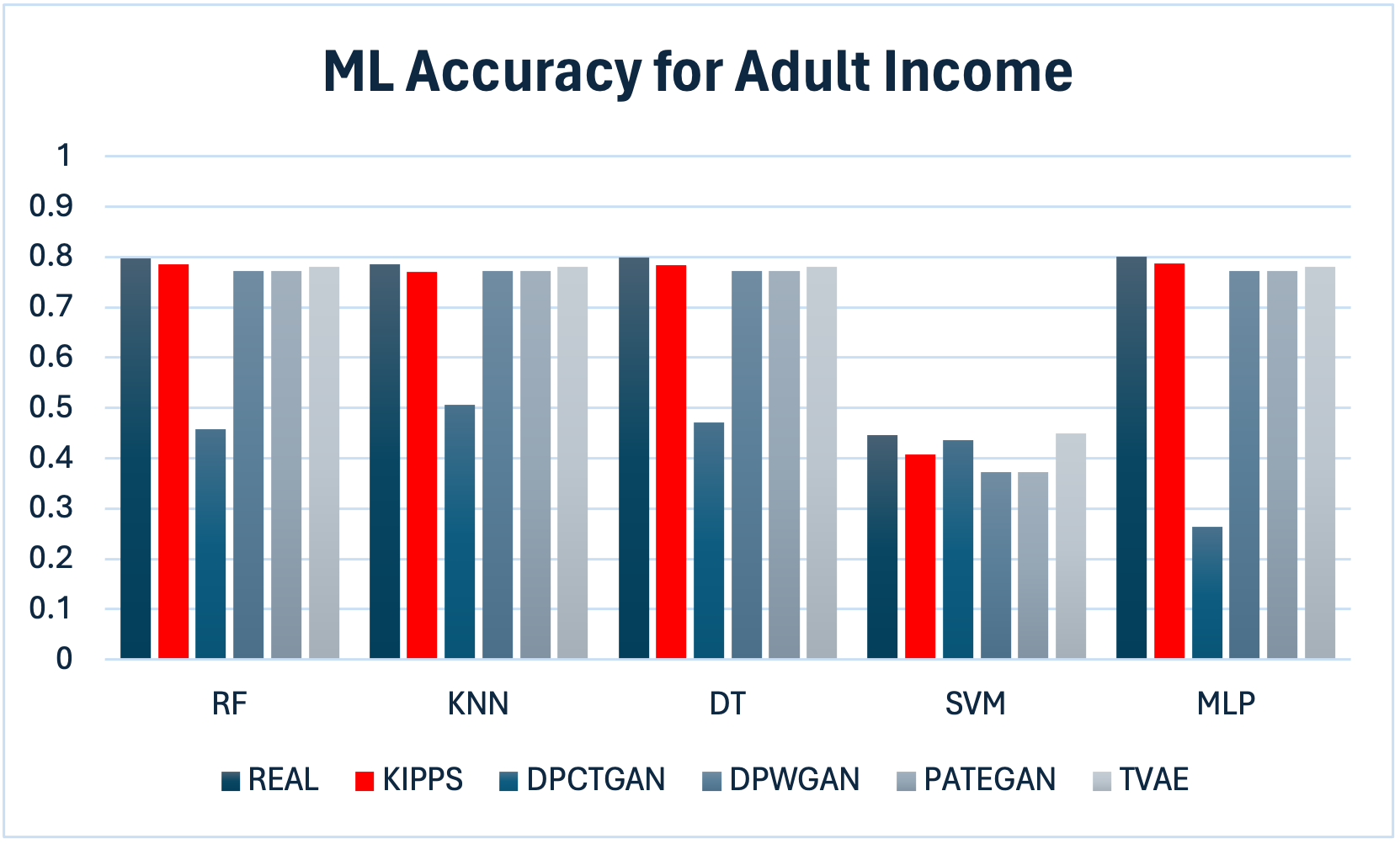}
  \label{}
\subcaption[]{Accuracy for Adult Income dataset}
\end{subfigure}
\begin{subfigure}{0.49\textwidth}
  \includegraphics[width=\textwidth]{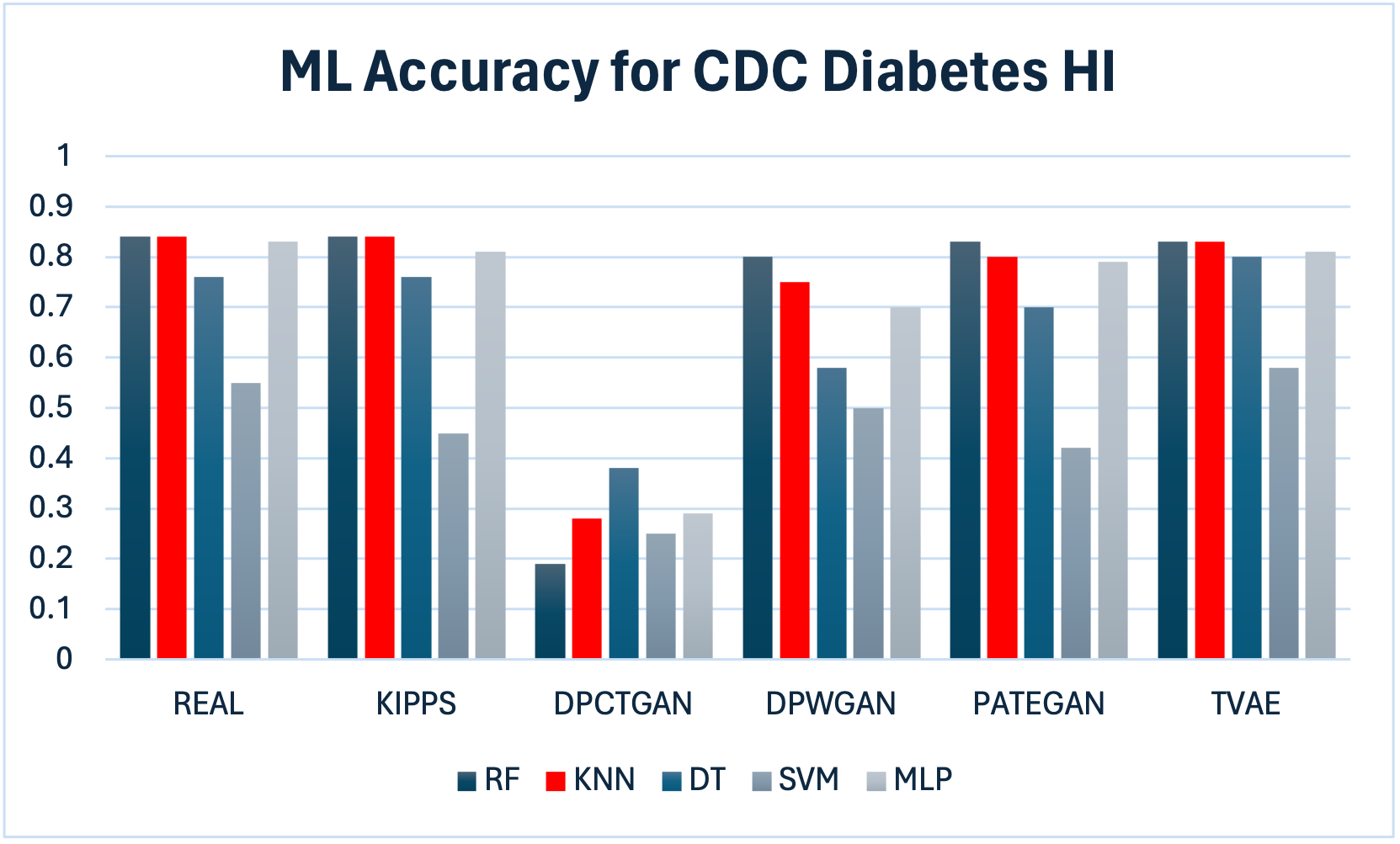}
  \label{}
  \subcaption[]{Accuracy for CDC Diabetes HI dataset}
\end{subfigure}
\begin{subfigure}{0.49\textwidth}
  \includegraphics[width=\textwidth]{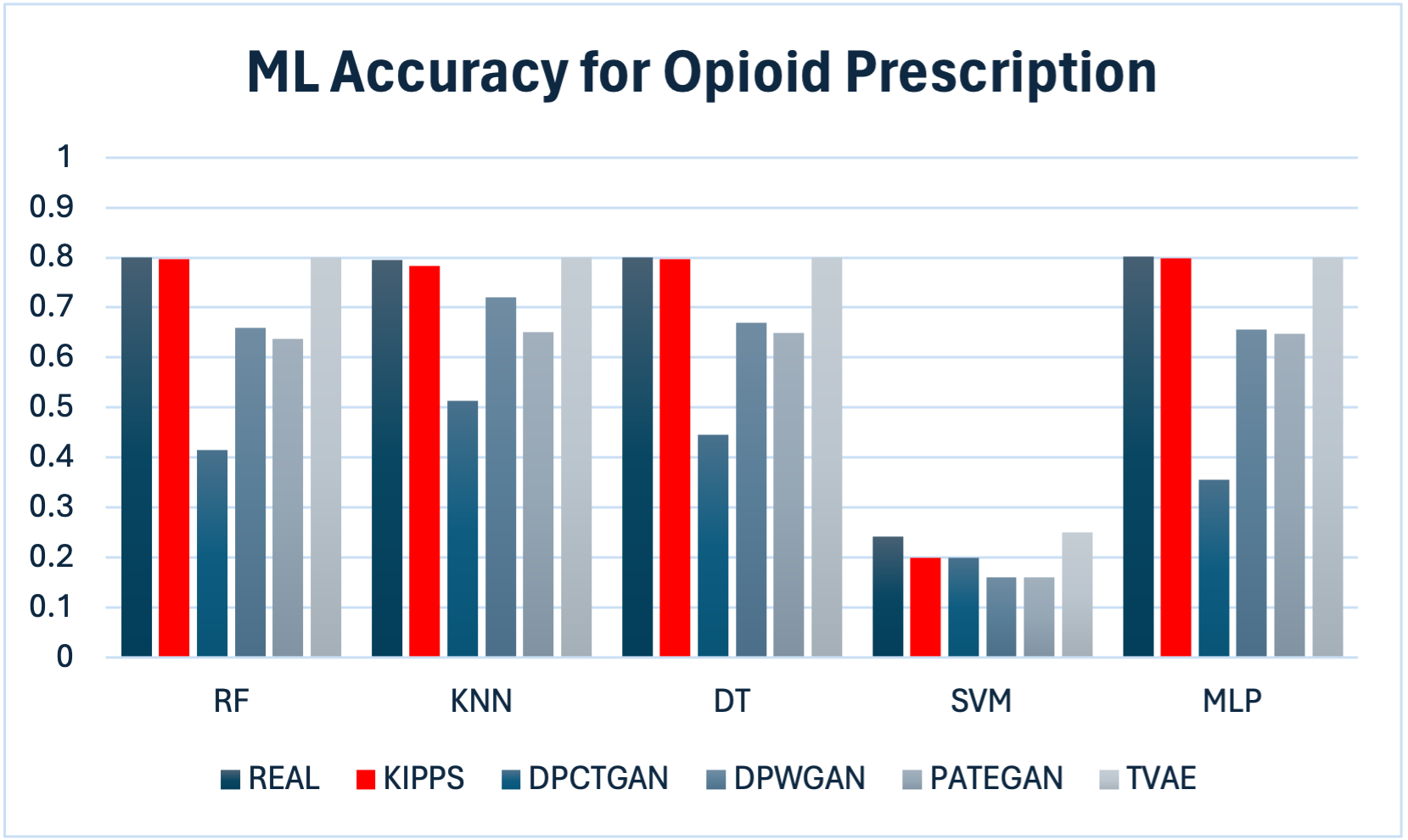}
  \label{}
  \subcaption[]{Accuracy for Opioid Prescription dataset}
\end{subfigure}
\begin{subfigure}{0.49\textwidth}
  \includegraphics[width=\textwidth]{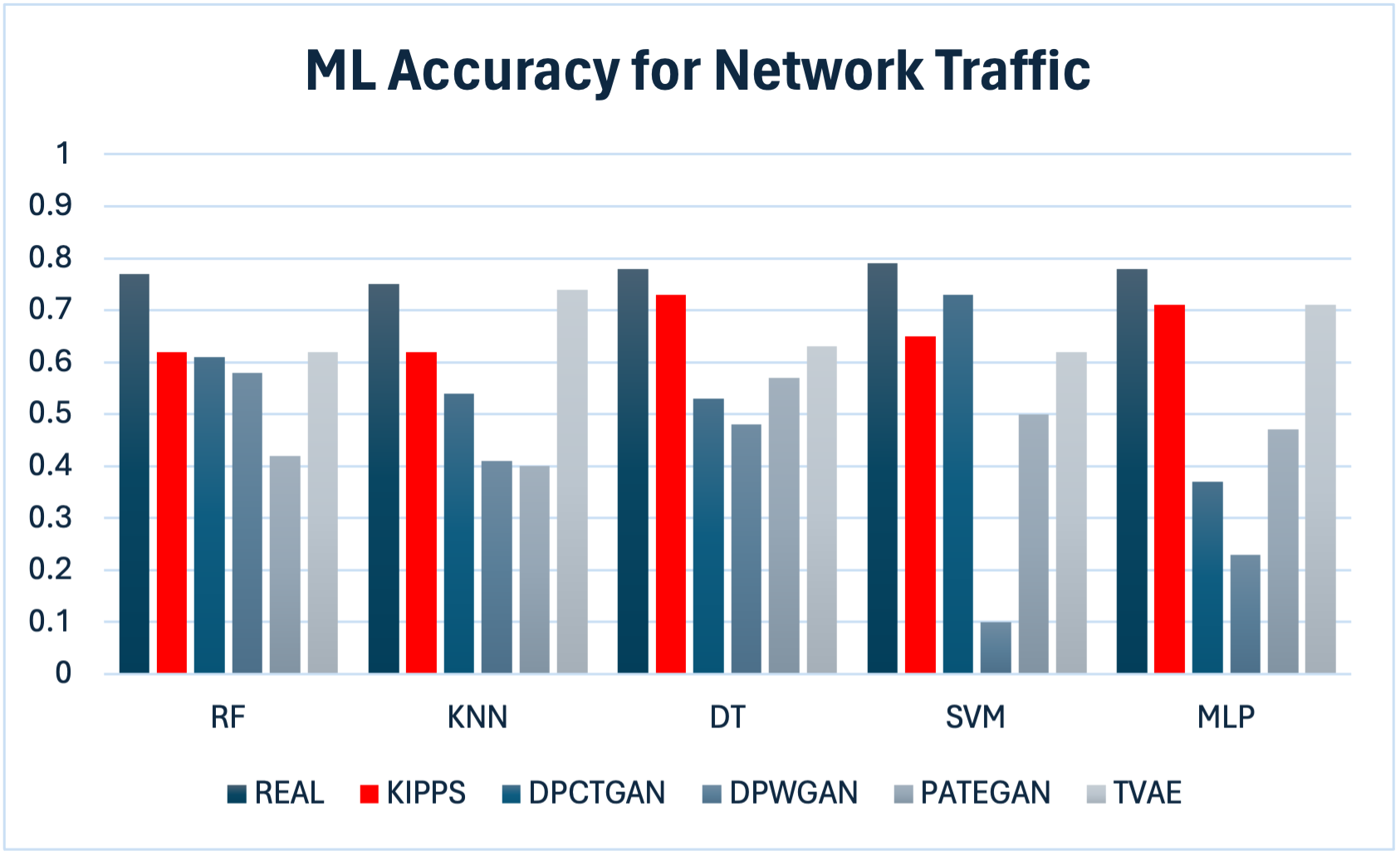}
  \label{}
  \subcaption[]{Accuracy for Network Traffic dataset}
\end{subfigure}
\begin{subfigure}{0.49\textwidth}
  \includegraphics[width=\textwidth]{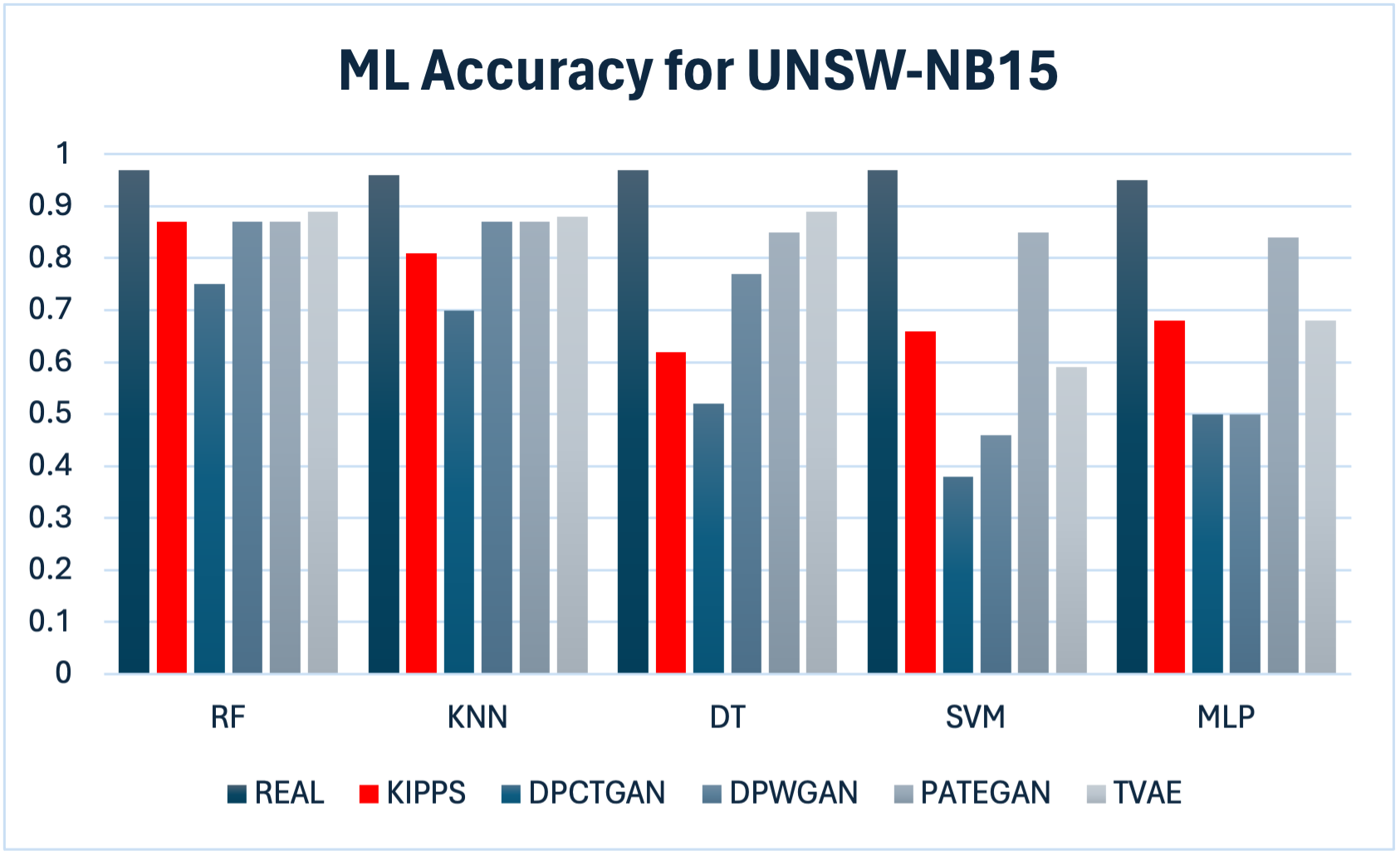}
  \label{}
  \subcaption[]{Accuracy for UNSW-NB15 dataset}
\end{subfigure}
\begin{subfigure}{0.49\textwidth}
  \includegraphics[width=\textwidth]{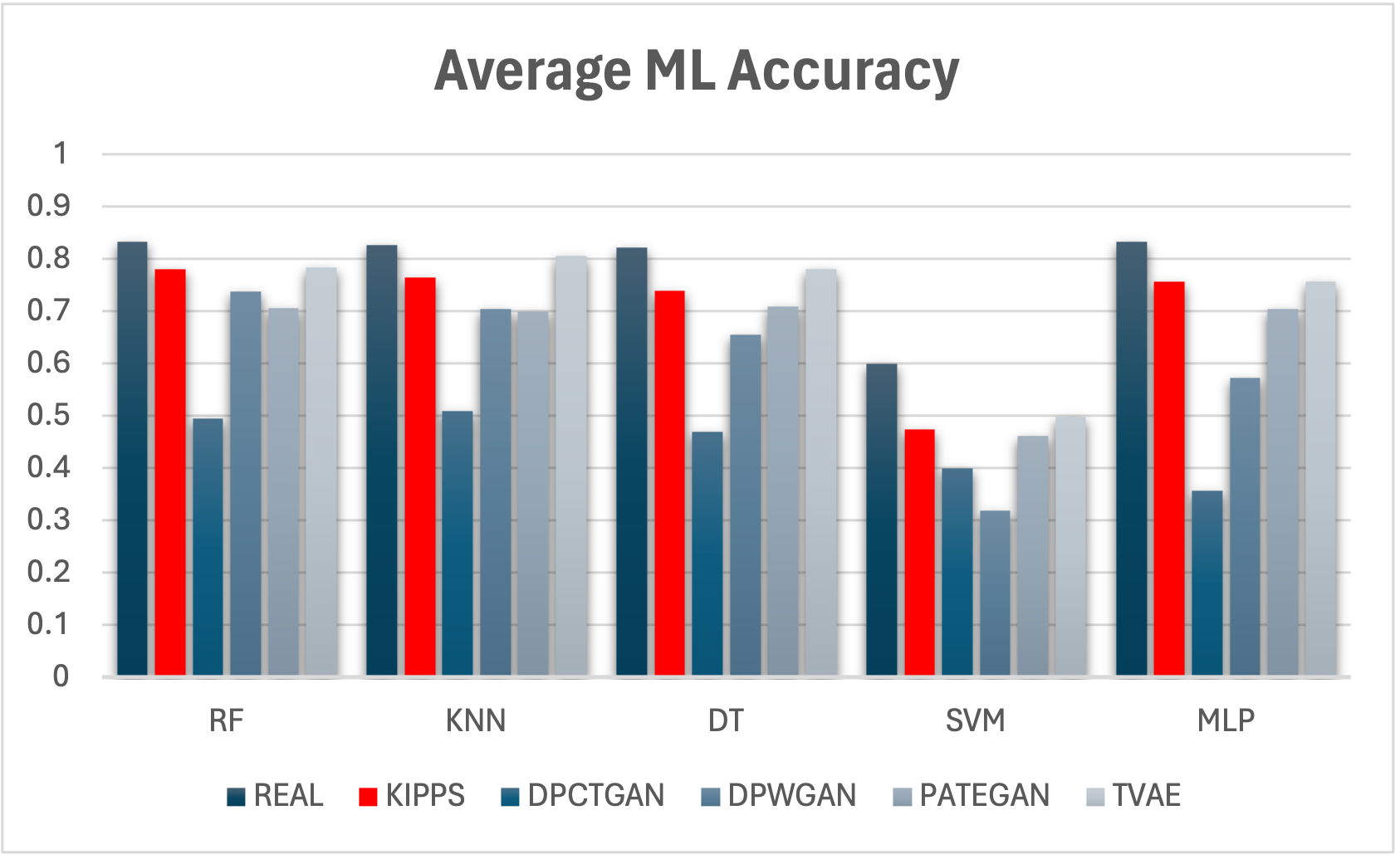}
  \label{}
  \subcaption[]{Average ML Accuracy}
\end{subfigure}
\caption{Comparison of Accuracy for Machine Learning Classifiers trained on real and synthetically generated data}
\label{fig:ml}
\end{figure*}%

\section{Experimental Results}
To validate our approach, we compare it with other state-of-the-art generative models for tabular data generation: DPCTGAN \cite{rosenblatt2020differentially}, DPWGAN \cite{xie2018differentially}, PATEGAN \cite{jordon2018pate}, and TVAE \cite{xu2019modeling}. To show the generalizability of the model we provide the results for datasets across different domains, such as healthcare, finance, and more as described in Section \ref{sec_datasets}. Experimental results are provided for synthetic data generated from various tabular datasets to demonstrate the robustness and applicability of our approach.

\subsection{Summary of Results}
\Algoname demonstrates a robust balance between high similarity and utility metrics while maintaining strong privacy protection, making it a favorable choice for synthetic data generation. In fidelity evaluations, \Algoname exhibits high predictive accuracy with a PMSE of 0.24 and better distributional similarity to real data, evidenced by higher p-values for $\chi^2$ and KS tests compared to other models. Utility results show \Algoname’s scores closely match those of real data across various machine learning algorithms, such as achieving 0.61 in Random Forest, which is nearly identical to the real data score of 0.62. Privacy evaluations reveal \Algoname's effectiveness in protecting against membership inference and attribute inference attacks, with average scores of 0.47 and 0.40 respectively, indicating strong privacy resilience. Overall, \Algoname excels in the privacy-utility tradeoff, delivering high-quality synthetic data while safeguarding sensitive information. The detailed results of each metrics is described in subsequent sections.

\begin{figure}
    \centering
    \includegraphics[width=\columnwidth]
    {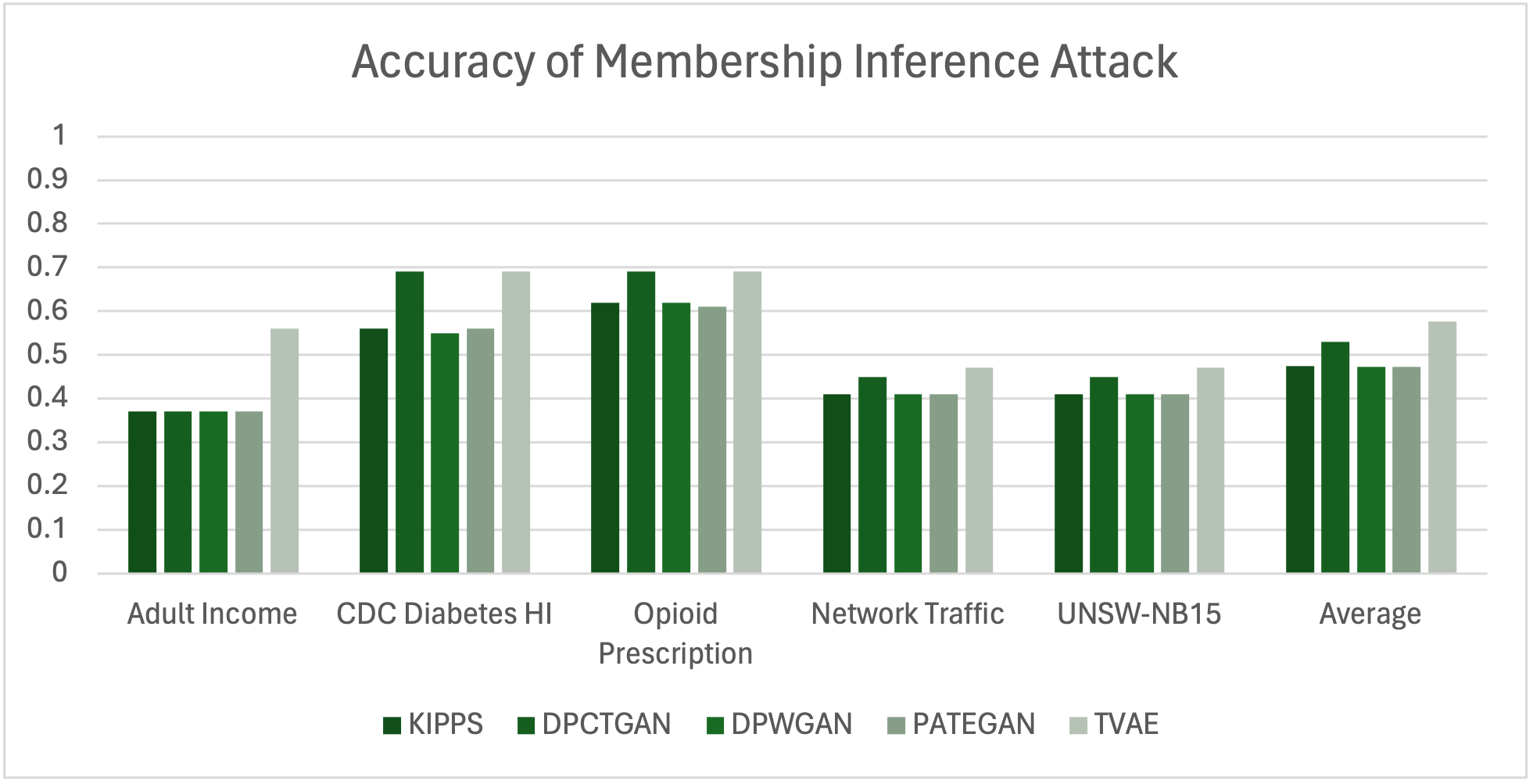}
    \caption{Comparison of mean accuracy for Membership Inference Attack}
    \label{fig:mia}
\end{figure}

\begin{figure}
    \centering
    \includegraphics[width=\columnwidth]
    {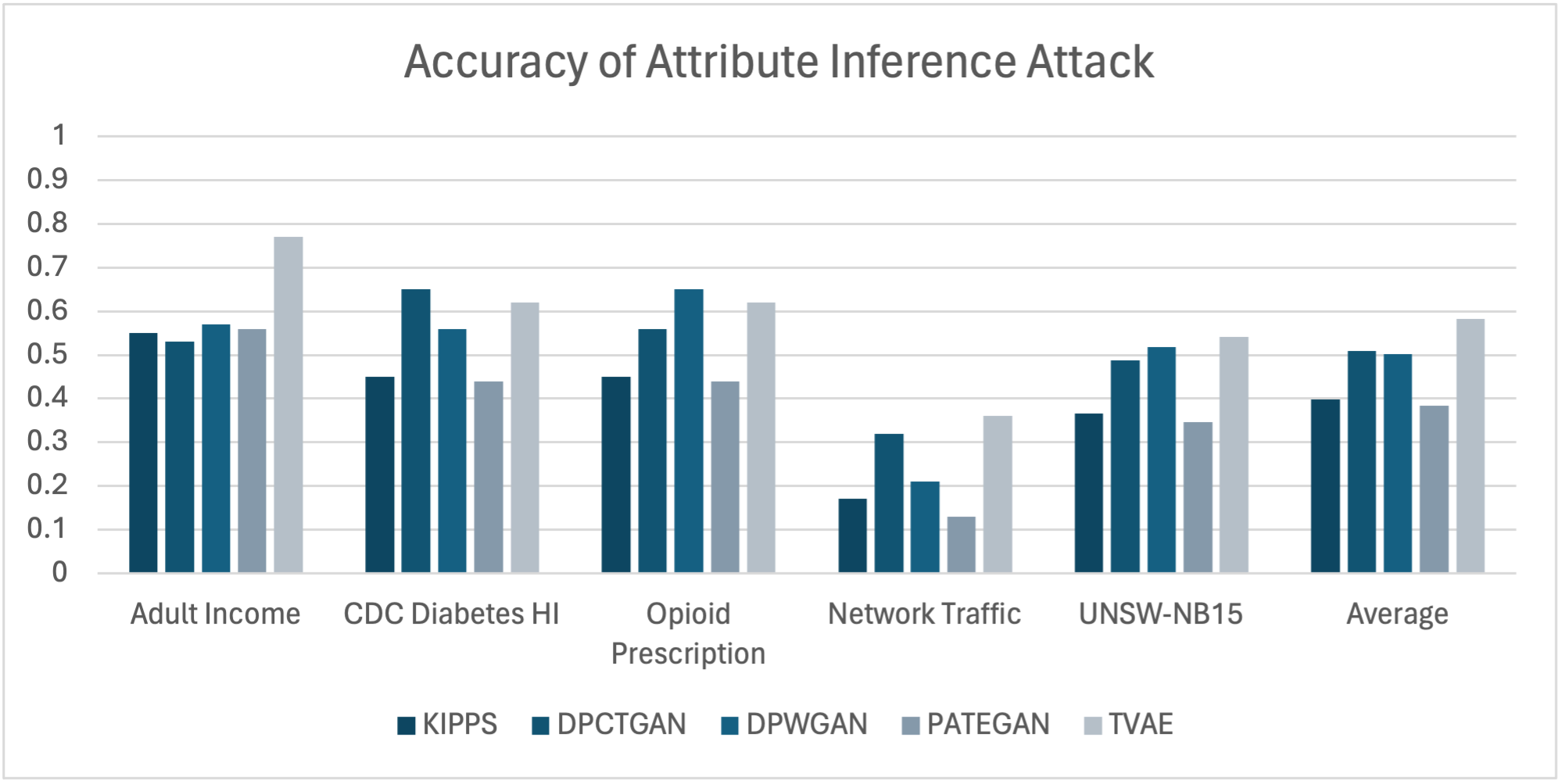}
    \caption{Comparison of mean accuracy for Attribute Inference Attack}
    \label{fig:AIA}
\end{figure}

\subsection{Fidelity Results}
The results for fidelity evaluation indicate that \Algoname performs favorably in several metrics compared to the other models. The results have been presented in Table \ref{tab:fidelity}. Here are the key points highlighting the strengths of \Algoname:

\begin{enumerate}
    \item Propensity Score: \Algoname has a PMSE of 0.24 for Predictive Propensity Score Accuracy, which is among the lowest, indicating high predictive accuracy.
    \item Distributional Distance: \Algoname achieves a high p-value for $\chi^2$ value of 0.08, significantly higher than DPCTGAN (0), DPWGAN (0.02), and PATEGAN (0.04). This suggests \Algoname fits the observed data better than most other models. \Algoname has a p-value for KS of 0.06, which, while equal to TVAE, is higher than DPCTGAN and PATEGAN (both at 0.02) and DPWGAN (0.01), indicating a better distributional similarity to the real data compared to these models.
    \item Regression Metrics: \Algoname has a high confidence interval overlap of 5.688, tied with PATEGAN, showing that it preserves the statistical properties of the original data effectively. \Algoname has a standardized difference of 0.45, which is lower than DPWGAN and PATEGAN (both at 0.5) and significantly lower than TVAE (9.74), suggesting better balance between data utility and privacy.
\end{enumerate}

\subsection{Utility Results}
Figure \ref{fig:ml} compares the performance of various models across multiple datasets and machine learning algorithms.The results show that \Algoname consistently delivers competitive machine learning accuracy across various algorithms, closely following original data. For RF, KNN, and DT, \Algoname achieves the second-highest accuracy scores of 0.7797, 0.76488, and 0.7382 respectively, just behind original data. In SVM and MLP, \Algoname also performs well with scores of 0.47358 and 0.75692, making it a robust synthetic data generation method. Overall, \Algoname demonstrates high and stable performance across different ML models, indicating its effectiveness in generating high-quality synthetic data that closely rivals original data. For the Adult Income Dataset \Algoname  achieves an average score of 0.706, which is only slightly lower than the real data's 0.726 and higher than DPCTGAN (0.427), DPWGAN (0.692), and PATEGAN (0.692). In the CDC Diabetes Health Indicator dataset, \Algoname performs robustly, matching the original data in RF and KNN with an accuracy of 0.84 and closely following in DT (0.76), SVM (0.45), and MLP (0.81). For the Opioid Prescription Dataset, \Algoname has an average score of 0.675, which is very close to the real data's 0.687 and higher than DPCTGAN (0.385), DPWGAN (0.572), and PATEGAN (0.55). In the Network Traffic dataset, \Algoname shows strong performance with competitive results in DT (0.73) and MLP (0.71), and maintains accuracy in RF (0.62) and KNN (0.62). In the UNSW-NB15 dataset, \Algoname shows commendable performance, particularly in RF (0.87), KNN (0.81), and MLP (0.68), demonstrating one of the top performancs in generating high-quality synthetic data.

\subsection{Privacy Results}
The results for the privacy resilience metric, which measures the accuracy of membership inference attacks, highlight the performance of various models trained with synthetic data. Figure \ref{fig:mia} demonstrates the accuracy score of membership inference attack against synthetic data generation models. In this context, lower scores indicate stronger privacy protection. The \Algoname algorithm achieved an average privacy resilience score of 0.47, demonstrating strong resilience against inference attacks. This is compared to DPCTGAN with a higher average score of 0.53, indicating weaker privacy protection, and TVAE with the highest score of 0.57. The results for the attribute inference attack accuracy, which measures the ability to deduce unknown attributes of individuals based on other known attributes, underscore the privacy protection afforded by various models trained with synthetic data. Here, lower accuracy values indicate stronger privacy resilience. The \Algoname algorithm achieved an average score of 0.40, as demonstrated in Figure \ref{fig:AIA}. This performance is superior to DPCTGAN, DPWGAN, and TVAE, which have higher average scores of 0.51, 0.50, and 0.58 respectively, suggesting weaker privacy protection. \Algoname performed exceptionally well on the Network Traffic dataset with a score of 0.17, significantly lower than the scores of DPCTGAN, DPWGAN, and TVAE. These results indicate that \Algoname effectively conceals sensitive attributes, ensuring the synthetic data can be used safely without revealing sensitive details about individuals in the original dataset.

\section{CONCLUSION}
With the increasing demand for large datasets, privacy and confidentiality concerns have become paramount. Strict regulations or prohibitions on data sharing in many fields necessitate robust privacy-preserving measures. Synthetic data generation offers a scalable solution, allowing organizations to collaborate on research and analysis without compromising sensitive information. The proposed \Algoname model enhances privacy-preserving synthetic data generation by incorporating domain knowledge through Knowledge Graphs, addressing challenges related to data diversity, complexity, and domain-specific constraints. \Algoname effectively balances high similarity and utility metrics with low privacy attack accuracy, making it an excellent choice for synthetic data generation. It demonstrates high predictive accuracy, better distributional similarity, and maintains strong privacy protection across multiple datasets. Evaluations show \Algoname’s scores closely match those of real data in various machine learning algorithms, proving its reliability. 

Future research can explore the integration of additional domain-specific knowledge to further improve the realism and utility of synthetic data. Investigating advanced differential privacy techniques may enhance privacy guarantees without compromising data utility. Additionally, expanding the application of \Algoname to other domains, such as finance and social sciences, could validate its versatility and effectiveness. Further studies could also focus on optimizing \Algoname for real-time data generation and exploring its performance in federated learning environments. Continuous refinement of the model to handle more complex and higher-dimensional data will be crucial in keeping pace with the evolving demands for privacy-preserving synthetic data.

\bibliographystyle{IEEEtran}
\bibliography{refs}

\end{document}